\documentclass[english, letterpaper, 10 pt, conference]{ieeeconf}
\usepackage[T1]{fontenc}
\usepackage[latin9]{inputenc}
\usepackage{amsmath}
\usepackage{cite}
\usepackage[caption=false]{subfig}
\usepackage{dsfont}
\usepackage{bm}
\usepackage{url}

\IEEEoverridecommandlockouts
\makeatletter

\newtheorem{defn}{Definition}

\usepackage{graphicx} 
\usepackage[binary-units=true]{siunitx}

\newcounter{tagnumb}
\newcommand\tageq[1]{\tag{#1\thetagnumb} \stepcounter{tagnumb}}

\makeatother

\usepackage{babel}

\title{Time-optimal Coordination of Mobile Robots along Specified Paths}

\author{ Florent Altch\'e$^{1,2}$, Xiangjun Qian$^{1}$ and Arnaud de La Fortelle$^{1}$%
\thanks{$^{1}$ MINES ParisTech, PSL Research University, Centre for robotics, 60 Bd St Michel 75006 Paris, France {\tt\small [florent.altche, xiangjun.qian, arnaud.de\_la\_fortelle] @mines-paristech.fr}}
\thanks{$^{2}$ \'Ecole des Ponts ParisTech, Cit\'e Descartes, 6-8 Av Blaise Pascal, 77455 Champs-sur-Marne, France}%
\thanks{This work has received support from the European FP7 project AutoNet2030 (Grant Agreement NO. 610542).}
}

\begin{document}

\maketitle
\thispagestyle{empty}
\pagestyle{empty}

\begin{abstract} In this paper, we address the problem of  time-optimal coordination of mobile robots under kinodynamic constraints along specified paths. We propose a novel approach based on time discretization that leads to a mixed-integer linear programming (MILP) formulation. This problem can be solved using general-purpose MILP solvers in a reasonable time, resulting in a resolution-optimal solution. Moreover, unlike previous work found in the literature, our formulation allows an exact linear modeling (up to the discretization resolution) of second-order dynamic constraints. Extensive simulations are performed to demonstrate the effectiveness of our approach.
\end{abstract}

\section{Introduction} The deployment of autonomous mobile robots is expected to bring major benefits in many applications, and their number is likely to grow dramatically in the next decades. Therefore, the need to coordinate these robots, which means finding a way for each of them to reach its target without colliding with another robot, will become increasingly important. A vast literature on motion planning, which generalizes the problem of coordination, already exists (see\textit{ e.g.}~\cite{latombe2012robot}).
 
We consider the problem of optimally coordinating multiple robots along specified paths with variable speed under kinodynamic constraints. The simplifying fixed path assumption is notably suited for structured environments, for instance traffic intersections or warehouses, where robots are generally bound to navigate inside lanes or aisles. Various methods exist to quickly find a \textit{feasible} solution to this problem, and many of these (see \textit{e.g.}~\cite{Simeon2002}) make use of the so-called coordination (or configuration) space introduced in~\cite{Lozano-Perez1983}. However, the \textit{optimal} coordination problem is known to be NP-hard~\cite{Dasler2015}, even in the absence of dynamic constraints, and few methods exist to provide a good solution with a high number of robots. As shown in~\cite{Gregoire2014}, this algorithmic complexity stems from the implicit decision of choosing which of any two potentially colliding robots should be the first to pass, making the problem inherently combinatorial with complexity scaling as high as $2^{N(N-1)/2}$ for $N$ robots. Exhaustive enumeration could be used for small instances (see \textit{e.g.}~\cite{Chitsaz2004,Murgovski2015}), but would be difficult to scale up with a larger number of robots.
 
A possible way of handling this complexity is to prune the corresponding decision tree by removing provably non-optimal branches. Mixed-integer programming (MIP) is a widely-used framework that allows efficient handling of such combinatorial problems. General MIP problems involving arbitrary functions are very hard, but good techniques exist for a subclass of these problems, called mixed-integer second-order cone programming, or MISOCP~\cite{Alizadeh2003}. In these problems, a convex quadratic objective function is minimized with quadratic positive semi-definite or affine constraints. A better-known subclass of MISOCP is mixed-integer linear programming (MILP), where the objective function and constraints are linear. These techniques have already been applied to trajectory planning in general, and to the coordination problem in particular.
 
In~\cite{Richards2002}, a MILP formulation is used to compute fuel-optimal trajectories for multiple spacecrafts, while avoiding collisions and exhaust gases from other crafts. However, this model is inherently different from ours as paths are not specified in advance in this formulation, leading to a much higher computational complexity. On the opposite end of the spectrum, Wang et al.~\cite{Wang2011} use MILP to find an optimal velocity profile under piecewise-constant dynamic constraints for a train on a fixed path, but do not consider conflicts with other trains.

In~\cite{Cafieri2014}, a mixed-integer nonlinear program is used  to ensure safe separation of aircrafts evolving along fixed paths, with minimal deviation from an original flight plan. Even though dynamic constraints are not considered, nonlinearities stemming from the euclidean distance constraints render the problem very hard to solve in reasonable time, with a practical limit of $4$ aircrafts.
 
Peng and Akella~\cite{Peng2005} consider a problem almost identical to ours. First, they propose to discretize robot paths into ``collision-free'' and ``conflicting'' segments. In a second step, by identifying the time instants when robots can enter and exit collision segments, they formulate collision avoidance constraints for every pair of robots into a nonlinear mixed-integer problem. However, the dynamic constraints are non-convex in this formulation, rendering the problem unsuitable for general-purpose solvers. To handle this difficulty, additional constraints are introduced to transform the problem into two linear subproblems that are solved as MILPs and give an upper and lower bound for the solution, but the actual optimal value is only approximated. 

The problem of non-convexity of the dynamic constraints is also encountered in~\cite{Murgovski2015}, where authors propose a convexification method using linearization along a reference speed. However, this approximation underestimates the maximum and overestimates the minimum acceleration, leading to a sub-optimal solution.

Our main contribution in this paper is the introduction of a new formulation of the optimal coordination problem as a MILP by discretizing over time instead of space, and using bounding polygons to transcribe safety requirements as linear constraints. This formulation allows an exact linear modeling of second-order dynamics up to the discretization resolution; solving this problem gives optimal velocity profiles for the robots under these constraints. For illustration purposes, we use the mean sojourn time as the minimization objective, but other linear optimization criteria could also be chosen.
 
The paper is articulated as follows. In Section~\ref{sec:System-modeling}, we describe the modeling of the system of robots and the general coordination problem, and we give some theoretical insights on the problem from previous work in Section~\ref{seq:theory}. In Section~\ref{sec:Minimum-delay-coordination}, we present our formulation of the time-optimal coordination problem as a MILP problem. This formulation is then validated using computer simulation on the example of automated vehicles in an intersection, as presented in Section~\ref{sec:Simulation-results}. Finally, Section~\ref{sec:Conclusion} concludes the study.

\section{Problem statement\label{sec:System-modeling}}
\begin{figure}
\centering\includegraphics{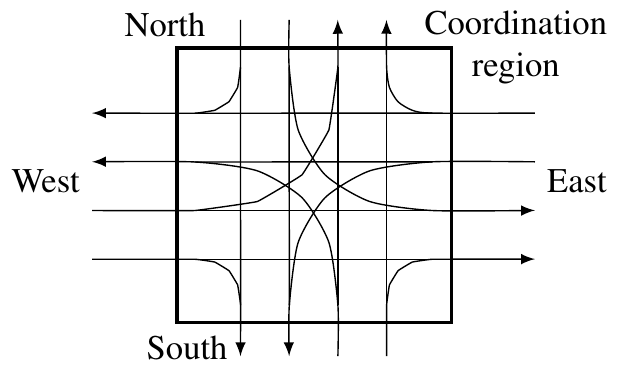}

\caption{\label{fig:Example-coord-region}Example of paths inside and outside
the coordination region for a two-lane road intersection.\vspace*{-0.5cm}}
\end{figure}

We consider a set $\mathcal{N}$ of $N$ robots evolving on predetermined paths, and we assume that coordination between robots is only needed inside a bounded region, which we call the \textit{coordination region}. In the case of automated driving, for instance, coordination is mostly needed in the middle of road intersections, as illustrated in Fig.~\ref{fig:Example-coord-region}, while vehicles only need to keep a safe distance with the vehicle in front of them when they are not in the intersection. In this example, the coordination region would be chosen as the center of the intersection, including a portion of the roads leading to, and exiting from this center. Note that the predefined path assumption allows to only consider longitudinal movement of the robots, thus reducing the computational complexity, and is classical in robots coordination problems in a constrained environment~\cite{Peng2005,Pecora2012}.

Each robot $i \in \mathcal N$ is supposed to follow a predetermined path $\gamma_{i}$ inside the coordination region, so that we need only consider the longitudinal comportment of the robots. The dynamics of robot $i$ are described as a double integrator:
\begin{equation}
(\dot{s}_i, \dot v_i) = (v_i, a_i) 
\end{equation} 
where $s_{i}$ is the curvilinear position of robot $i$ along its path, $v_i$ its longitudinal velocity and $a_i$ its acceleration.

The origin of $s_{i}$ is chosen so that $s_{i}=0$ when the front of the robot enters the coordination region, and $s_{i}=s_{i}^{out}>0$ when it fully exits the coordination region. $s_i$ can therefore be interpreted as the distance traveled by robot $i$ inside the coordination region. The velocity is assumed to be non-negative and bounded, such that $v_{i} \in [0,\overline{v}_i]$. The state of robot $i$ is noted $x_{i}=(s_{i},v_{i})$; we call \textit{trajectory} of robot $i$ the (continuous) function mapping a given time $t$ to the state of $i$ at time $t$, noted $x_{i}(t) = (s_i(t), v_i(t))$.  Boldface $\bm{x}$ denotes the vector $(x_1, x_2, ...)$, representing the state of the multi-robot system and $\bm{x}(t)$ is the system trajectory.
To account for the dynamic constraints on the robots, we assume that the longitudinal acceleration $a_i$ of robot $i$ is bounded to an interval $\left[\underline{a_{i}},\overline a_{i}\right]$ with $\underline a_{i}<0<\overline a_{i}$.

We assume that robot $i$ enters the coordination region at time $t_{i}^{in}$ with speed $v^{in}_i \in [0,\overline{v}_i]$, and is required to leave the coordination region with speed $v^{out}_i$; we let $t_{i}^{out}$ denote the corresponding exit time. Note that $v^{out}_i$ should be properly chosen to avoid collisions outside of the coordination region, for instance when a fast vehicle exits after a slower one.

For a pair of distinct robots $i$ and $j$, we call \textit{collision set} between $i$ and $j$, noted $\mathcal{C}_{ij}\subset\left[0,s_{i}^{out}\right]\times\left[0,s_{j}^{out}\right]$ the set of positions $(s_{i},s_{j})$ inside the coordination region where $i$ and $j$ would collide; we say that $i$ and $j$ are \textit{conflicting} if $\mathcal{C}_{ij}\neq\emptyset$.

\begin{figure}[h]\centering
\subfloat[Following paths]{\includegraphics[width=0.23\textwidth]{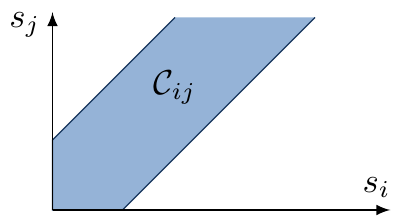}}
\subfloat[\label{fig:Crossing-paths}Crossing paths]{\includegraphics[width=0.23\textwidth]{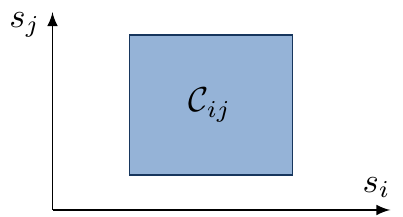}}

\subfloat[\label{fig:Merging-paths}Merging paths]{\includegraphics[width=0.23\textwidth]{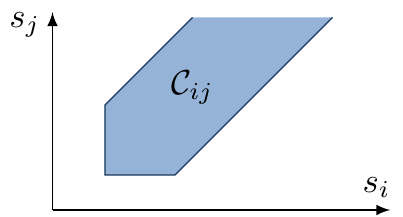}}
\subfloat[Diverging paths]{\includegraphics[width=0.23\textwidth]{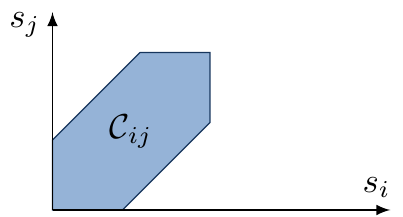}}
\caption{\label{fig:Collision-spaces}Possible cases for the collision set between two conflicting rectangular robots $i$ and $j$ (assuming paths intersect orthogonally). The edges of $\mathcal C_{ij}$ are straight lines with equations $s_i = A$, $s_j = B$ or $|s_i-s_j| = C$ where $A, B, C$ are constants.\vspace*{-0.2cm}}
\end{figure}

To simplify the rest of the presentation, we assume that the collision set between two robots is connected, \textit{i.e.} we exclude the case of non-conflicting segments between two conflict segments, for instance when two paths intersect multiple times. If this is not the case, the presented results still hold provided every connected component $\mathcal{C}_{ij}^{p}$ of $\mathcal{C}_{ij}$ is considered individually. Moreover, we approximate the exact collision set by a (minimal) bounding polygon with edges either parallel to the horizontal or vertical axis, or to the $s_{i}=s_{j}$ line. Under our hypotheses, four possible types of conflicts can exist: robots can follow one another throughout the coordination region (a), or have crossing (b), merging (c) or diverging (d) paths, as illustrated in Fig.~\ref{fig:Collision-spaces}. Note that cases (c) and (d) are not mutually exclusive, and paths can merge and then diverge.\vspace*{-0.5cm}

\begin{figure}[h]
\subfloat[Robots shape]{\includegraphics[height=3cm]{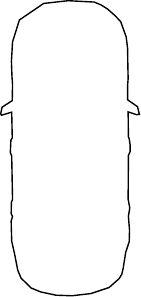}}
\hfill
\subfloat[\label{fig:Crossing-paths-exact}Crossing paths]{\includegraphics[height=3cm]{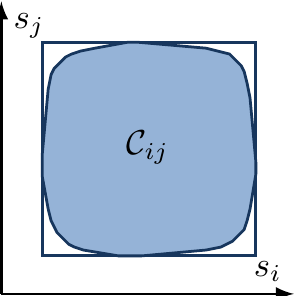}}
\hfill
\subfloat[\label{fig:Merging-paths-exact}Merging and diverging paths]{\includegraphics[height=3cm]{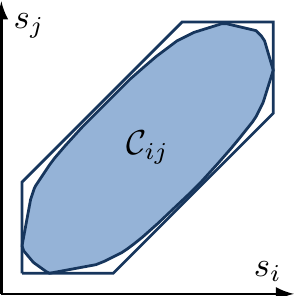}}
\caption{\label{fig:Exact-coll-reg}Exact shape of the collision region $\mathcal C_{ij}$ for (polygonal) car-like robots. The blue rectangles represent the corresponding bounding polygon.}
\end{figure}

The above bounding polygon approximation is well suited when robots have nearly rectangular shapes, and induces a negligible loss of optimality in this case, as illustrated in Fig. \ref{fig:Exact-coll-reg}. Note that arbitrary convex polygons can also be used, at the cost of introducing additional binary variables.

We now define the time-optimal coordination problem as:
\begin{defn}The \textit{time-optimal coordination problem} is that of finding the optimal system trajectory $\bm{x}^*(t)$ minimizing the mean sojourn time $\frac{1}{N} \sum_{i \in \mathcal{N}}(t^{out}_i - t^{in}_i)$, under the following constraints for every robot $i$:
\begin{itemize}
\item initial conditions: $x_{i}(t_{i}^{in})=(0,v_{i}^{in})$
\item kinematics: $\dot{s_{i}}=v_{i} \in [0, \overline v_i]$
\item dynamics: $\dot{v_{i}}=a_{i} \in [\underline{a}_i, \overline{a}_i]$
\item safety: for all $t\geq t_{i}^{in}$ and for every robot
$j\neq i$, $(s_{i}(t),s_{j}(t))\notin\mathcal{C}_{ij}$
\item liveness: there exists $t_{i}^{out}<+\infty$ with $s_{i}(t_{i}^{out})=s_{i}^{out}$
\item exit speed: $v_{i}(t_{i}^{out})=v^{out}_i$.
\end{itemize}
\end{defn}
Note that in the above definition, the initial condition $x_i(t_i^{in}) = (0, v_i^{in})$ corresponding to a fixed entry time can be replaced by a fixed initial state constraint $x_i(0) = (s_i^{0}, v_i^{0})$.

\section{Theoretical analysis\label{seq:theory}}
For an arbitrary pair of robots with non-empty collision set, one necessarily passes before the other to avoid collisions. For two conflicting robots $i$ and $j$ and a given collision-free system trajectory, we say that $i$ has priority over $j$ if $i$ goes before $j$, and we note $i \succ j$ in this case. The collection of all priorities for all pairs of robots can be encoded as a priority graph where robots are the nodes and priorities are directed edges. It has been shown in \cite{Gregoire2014} that such priority graphs can be bijectively mapped to homotopy classes of trajectories for the multi-robot system. For a given continuous optimization criterion, there exists at least one optimal trajectory in any non-empty homotopy class. The global optimal trajectory $\bm{x}^*(t)$ can then be found by enumerating all optimal trajectories for all priority graphs.

Therefore, the time-optimal coordination problem has both a discrete (enumerating all priority graphs) and a continuous part (optimizing the trajectories of every robots under assigned priorities, which is a continuous optimal control problem). The discrete part of the problem is combinatorial since, for $N$ robots inducing $p$ pairs of conflicts, there are up to $2^{p}$ possible priority graphs. In general, $p$ can be as high as $\frac 1 2 N(N-1)$; however, many of these graphs can be discarded for poor performance or for being incompatible with the robots entry times. For this reason, \textit{branch-and-bound} algorithms seem particularly well-suited to our problem as they are designed to find global optima without needing to explore the whole decision tree.
 
\section{Time-optimal coordination\label{sec:Minimum-delay-coordination}}
Using the above theoretical results, we formulate the time-optimal coordination problem as a mixed-integer \textit{linear} program (MILP) which can be solved by widely-available solvers using branch-and-bound techniques.

\subsection{Completed Collision Set}\label{seq:completed-coll}
For conflicting robots $i$ and $j$, we call \textit{completed collision set} $\mathcal{C}_{i\succ j}$ the set of configurations $(s_{i}, s_{j})$ leading to a collision or a violation of priority $i\succ j$; mathematically, $\mathcal{C}_{i\succ j} = \mathcal{C}_{ij} + (\mathds R_- \times \mathds R_+)$. Since we have approximated each collision set $\mathcal{C}_{ij}$ by a minimal bounding polygon with edges parallel to the coordinate axes or to the $s_i = s_j$ lines, the completed set $\mathcal{C}_{i\succ j}$ is also a polygon with the same properties. Note that if $\mathcal{C}_{ij}$ is not connected, a completed collision set has to be defined for each connected component.

\begin{figure}
	\subfloat[\label{fig:Merging-paths-completed}Merging paths]{\includegraphics[width=3.4cm]{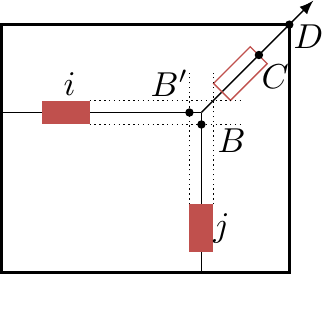}} \hfill 
	\subfloat[\label{fig:Completed-diagram}Corresponding collision set]{\includegraphics[width=5.2cm]{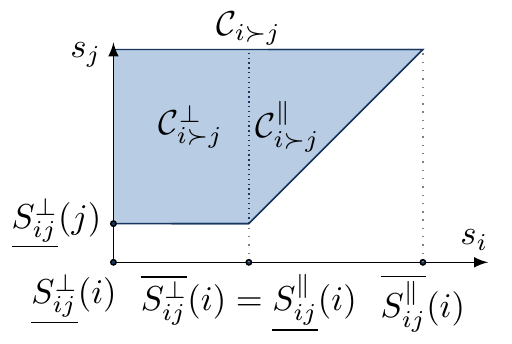}}
	\caption{\label{fig:Completed-obstacle}Illustration of the completed collision set $\mathcal{C}_{i\succ j}$ in the merging case, and the corresponding values of $\underline S_{ij}^{\perp}$,
		$\overline S_{ij}^{\perp}(i)$, $\underline S_{ij}^{\parallel}(i)$ and
		$\overline S_{ij}^{\parallel}(i)$. $s_j = \underline S_{ij}^\perp(j)$ when $j$ reaches $B$ in Fig.~\ref{fig:Merging-paths-completed}, and $s_i = \underline S_{ij}^\parallel(i)$ when $i$ reaches $C$. Similarly, $s_i = \underline S_{ji}^\perp(i)$ when $i$ reaches $B'$. Also note that the equation for the lower-right boundary of the collision set is $s_{i}\geq s_{j}+\underline S_{ij}^{\parallel}(i)-\underline S_{ij}^{\perp}(j)$.\vspace*{-0.5cm}}
\end{figure}

To formulate safety requirements as linear constraints, we first define a partition of $\mathcal{C}_{i\succ j}$ as $\mathcal{C}_{i\succ j}^{\parallel} \cup \mathcal{C}_{i\succ j}^{\perp}$. $\mathcal{C}_{i\succ j}^{\parallel}$ is the subset of $\mathcal{C}_{i\succ j}$ with boundary parallel to $s_i = s_j$, and $\mathcal{C}_{i\succ j}^{\perp}$ the subset with boundary parallel to the $s_i$ axis as illustrated for the merging case in Fig.~\ref{fig:Completed-obstacle}. Subset $\mathcal{C}_{i\succ j}^{\parallel}$ is the set of priority violations (or collisions) that could happen when $j$ should follow $i$, and $\mathcal{C}_{i\succ j}^{\perp}$ is that of violations (or collisions) that could happen when $j$ should wait for $i$ to pass. In what follows, we note $\underline S_{ij}^{\parallel}(\cdot)$ and $\overline{S}_{ij}^{\parallel}(\cdot)$ the lower and upper bounds of the projection of $\mathcal{C}_{i\succ j}^{\parallel}$ on the $s_{\cdot}$ axis (for $\cdot = i$ or $j$), and we define $\underline S_{ij}^{\perp}(\cdot)$ and $\overline S_{ij}^{\perp}(\cdot)$ similarly. However, to ensure those subsets have empty intersection, we exclude $\left\{\left(\overline S_{ij}^{\perp}(i),s_j\right) \ :\ s_j \in [0,s_j^{out}] \right\}$ from $\mathcal{C}_{i\succ j}^{\perp}$. If either $\mathcal{C}_{i\succ j}^{\parallel}$ or $\mathcal{C}_{i\succ j}^{\perp}$ is empty, we let the corresponding lower and upper bounds be equal to $0$; note that in this particular case, those subsets formally do not form a partition.

An important remark is that $\mathcal{C}_{i\succ j}^{\parallel}$ and $\mathcal{C}_{i\succ j}^{\perp}$ are both connected and left invariant by translation of a vector from $\{0\} \times \mathds R_+$. Therefore, with this decomposition, if $\underline S_{ij}^{\parallel}(i) \leq s_i \leq \overline S_{ij}^{\parallel}(i)$, condition $(s_{i}, s_{j}) \notin \mathcal{C}_{i\succ j}$ is equivalent to $s_j \leq s_i - a_{ij}$ where $a_{ij}$ is a constant corresponding to a following distance and a potential offset of curvilinear abscissa. If $\underline S_{ij}^{\perp}(i) \leq s_i < \overline S_{ij}^{\perp} (i)$, condition $(s_{i}, s_{j}) \notin \mathcal{C}_{i\succ j}$ is equivalent to $s_j \leq \underline S_{ij}^{\perp}(j)$. If $s_i$ is not in either of these intervals, collisions or priority violations cannot occur. Note that, if needed, a finer decomposition could be used to better approximate the exact shape of the collision set, by using bounding convex polygons with edges having slopes different than $0$ or $1$, and defining one subset of $\mathcal{C}_{i\succ j}$ per edge.

\subsection{Discretization} 
If priorities have been chosen, finding the best trajectories in continuous time and space is equivalent to a nonlinear time-optimal control problem in a non-convex space, which is usually difficult to solve. Several authors, among which \cite{Peng2005} and \cite{Murgovski2015}, have used spatial discretization to overcome this difficulty and formulate a simpler problem. In this setting, paths are divided into segments and the average speed $v_{avg}$ and occupancy time $t_{occ}$ for each robot in each segment are chosen as variables. One of the major issues arising from this discretization is that the second-order dynamics of the robot imply nonlinear (and non-convex) relations between the length $\ell$ of a segment and these two variables. Indeed, these constraints are written as $\ell=v_{avg}t_{occ}$ which translates to $\ell \leq v_{avg}t_{occ} \wedge \ell \geq  v_{avg}t_{occ}$; one of those inequalities defines a non-convex set in the $(v_{avg},t_{occ})$ plane that has to be approximated in order to be treated in most optimization frameworks, thus degrading solution quality.
 
Instead of using spatial segments, we propose a temporal discretization which, to the best of our knowledge, has not yet been used to solve the coordination problem with a fixed-paths assumption. In what follows, we note $\tau>0$ the fixed duration of a time step; for an integer $k\geq0$ and a robot $i$, we note $s_{i}^{k}=s_{i}(k\tau)$, and use similar notations for $v_{i}$. We note $K$ the maximum number of time steps in the computation.

\subsection{Variables}
 
Besides variables $s_{i}^{k}$ and $v_{i}^{k}$, we need to introduce a few supplementary variables, all of them binary. For two conflicting robots $i$ and $j$, we let $\pi_{ij}=1$ if and only if (iff) $i \succ j$, and for all time steps $k$ we define several variables $\varepsilon_{ij}$ that indicate if, at time step $k$, robots $i$ and $j$ have entered and/or exited $\mathcal C_{i\succ j}^\perp$ and $\mathcal C_{i\succ j}^\parallel$.
 
Specifically, we let $\varepsilon_{ij}^{\bullet,in}(i,k)=1$ iff $s_{i}^{k}\geq\underline S_{ij}^{\bullet}(i)$ and $\varepsilon_{ij}^{\bullet,out}(i,k)=1$ iff $s_{i}^{k}\geq\overline S_{ij}^{\bullet}(i)$, where $\bullet$ is either $\parallel$ or $\perp$, and we define similarly $\varepsilon_{ij}^{\bullet,in}(j,k)=1$ iff $s_{j}^{k}\geq\underline S_{ij}^{\bullet}(j)$ and $\varepsilon_{ij}^{\bullet,out}(j,k)=1$ iff $s_{j}^{k}\geq\overline S_{ij}^{\bullet}(j)$.
 
We also introduce, for every robot, the variables $\mu_{i}^{k}=1$ iff $s_{i}^{k}\geq0$ and $\sigma_{i}^{k}=1$ iff $s_{i}^{k}\geq s_{i}^{out}$, respectively indicating if the robot has entered and exited the coordination region at time step $k$.

\subsection{Objective function}
 
As stated earlier, we wish to minimize the average sojourn time, which is equivalent (since entry times are prescribed) to minimizing the average exit times. Since our formulation does not use time as a variable, we use
\begin{equation}
\mathcal{O} = \frac{1}{N}\sum_{\substack{i=1\dots N\\ k=0\dots K}}\sigma_{i}^{k}\label{eq:objective-function}
\end{equation} for our objective function. $\mathcal{O}$ is the average number of time steps spent after exiting the coordination region, which approximates (to the duration of a time step $\tau$) the total amount of time spent after exiting the coordination region, divided by $\tau$. More precisely, $\mathcal{O}=\frac{1}{N}\sum_{i=1\dots N}\left(K-k_{i}^{out}+1\right)$ where $k_{i}^{out}$ is the time step at which robot $i$ exits the coordination region, and thus maximizing $\mathcal{O}$ is equivalent to minimizing the average value of $k_{i}^{out}$.

\subsection{Constraints}
Many of the constraints needed for the coordination problem are conjunctions (noted $\wedge$), or logical implications ($\Rightarrow$); we use the binary variables introduced in the previous section as indicators, and use a ``big-$M$'' formulation~\cite{Richards2005} to enforce those constraints. In what follows, left-hand side terms are variables, and right-hand side terms are problem parameters.

\paragraph{Binary variables}
To ensure the additional binary variables are set to the correct value, the following helper conditions are enforced for all $0\leq k\leq K$, every robot $i\in \mathcal N$ and every robot $j\in \mathcal N$ conflicting with $i$:
\begin{align}
\mu_{i}^{k}=0 & \Rightarrow s_{i}^{k}\leq0\tageq{h}\label{eq:cstr-mu-inside}\\
\mu_{i}^{k}=1 & \Rightarrow s_{i}^{k}\geq0\tageq{h}\label{eq:cstr-mu-end}\\
\sigma_{i}^{k}=0 & \Rightarrow s_{i}^{k}\leq s_{i}^{out}\tageq{h}\label{eq:cstr-sigma-inside}\\
\sigma_{i}^{k}=1 & \Rightarrow s_{i}^{k}\geq s_{i}^{out}\tageq{h}\label{eq:cstr-sigma-end}\\
\pi_{ij}+&\pi_{ji} = 1 \tageq{h}\label{eq:cstr-pij-xor}
\end{align} Constraints similar to \eqref{eq:cstr-mu-inside}-\eqref{eq:cstr-mu-end} are used for $\varepsilon_{ij}^{\bullet,in}(i,k)$, $\varepsilon_{ij}^{\bullet,in}(j,k)$, $\varepsilon_{ij}^{\bullet,out}(i,k)$ and $\varepsilon_{ij}^{\bullet,out}(j,k)$, where $\bullet$ is either $\perp$ or $\parallel$. Constraint \eqref{eq:cstr-pij-xor} ensures that exactly one of the variables $\pi_{ij}$ and $\pi_{ji}$ is set to $1$. Note that strict inequalities cannot be enforced in a MILP framework; therefore, the above constraints do not specify the value of each indicator variables at its point of discontinuity. This limitation, however, has little effect on the solutions.
\setcounter{tagnumb}{0}

\paragraph{Initial and terminal values, bounds}
To account for the initial values of the variables and the different bounds on the problem, we use the following boundary constraints for all $0\leq k\leq K-1$ and each robot $i$:
\begin{align}
\mu_{i}^{k}=0\Rightarrow\, & v_{i}^{k+1}=v_{i}^{in}\tageq{b}\label{eq:cstr-init-speed}\\
\sigma_{i}^{k+1}=1\Rightarrow\, & v_{i}^{k}=v_i^{out}\tageq{b}\label{eq:cstr-final-speed}\\
s_{i}^{0}=\, & -v_{i}^{in}t_{i}^{in}\tageq{b}\label{eq:cstr-init-pos}\\ 
s_{i}^{K}\geq\, & s_{i}^{out}\tageq{b}\label{eq:cstr-final-pos}\\ 
v_{i}^{0}=\, & v_{i}^{in}\tageq{b}\label{eq:cstr-init-vel}\\ 
v_{i}^{k} \in\, & [0; \overline v_i] \tageq{b}\label{eq:cstr-minmax-vel}
\end{align}
\setcounter{tagnumb}{0} Conditions (\ref{eq:cstr-init-speed}), (\ref{eq:cstr-init-pos}) and (\ref{eq:cstr-init-vel}) enforce the initial constraint $x_{i}(t_{i}^{in})=(0,v_{i}^{in})$,  (\ref{eq:cstr-final-speed}) the final speed constraint, (\ref{eq:cstr-minmax-vel}) enforces the bounds on speed and (\ref{eq:cstr-final-pos}) ensures the liveness constraint, as all robots are required to exit in finite time.

\paragraph{Kinodynamic constraints}
We assume that robots use a constant acceleration during each time step. Under this assumption, we enforce the kinodynamic constraints using the conditions, for all $0\leq k\leq K-1$ and every robot $i$:
\begin{align}
\sigma_{i}^{k}=0\Rightarrow\ & s_{i}^{k+1}-s_{i}^{k}-\frac{1}{2} \tau \left(v_{i}^{k+1}+v_{i}^{k}\right)=0\tageq{k}\label{eq:cstr-kinematic-inside}\\
\sigma_{i}^{k}=0\Rightarrow\ & v_{i}^{k+1}-v_{i}^{k}\leq\overline a_{i}\tau\tageq{k}\label{eq:cstr-accel-ub}\\
\sigma_{i}^{k}=0\Rightarrow\ & v_{i}^{k+1}-v_{i}^{k}\geq\underline a_{i}\tau\tageq{k}\label{eq:cstr-accel-lb}
\end{align}
\setcounter{tagnumb}{0} Condition (\ref{eq:cstr-kinematic-inside}) enforces the kinematic constraints (under the constant acceleration assumption); (\ref{eq:cstr-accel-ub}) and (\ref{eq:cstr-accel-lb}) account for the dynamic constraints. Note that the constant acceleration hypothesis could be relaxed using third-order dynamics, or more general second-order cone programming, as
\begin{align}
  x_i^k - v_i^k \tau + \frac{\overline a_i \tau}{\underline a_i - \overline a_i} y_i^k  &- \frac{1}{2 (\underline a_i - \overline a_i) } \left. y_i^k \right.^2 \leq  \frac{\overline a_i\, \underline a_i \tau^2}{2 (\underline a_i-\overline a_i)} \nonumber \\
  x_i^k - v_i^k \tau + \frac{\underline a_i \tau}{\overline a_i - \underline a_i} y_i^k  &- \frac{1}{2 (\overline a_i - \underline a_i) } \left. y_i^k \right.^2 \geq  \frac{\overline a_i\, \underline a_i \tau^2}{2 (\overline a_i-\underline a_i)}  \nonumber
\end{align} 
where $x_i^k = s_i^{k+1} - s_i^k$ and $y_i^k = v_i^{k+1} - v_i^k$. A justification for this extension can be found in~\cite{Johnson2012} which shows that these constraints exactly describe the set of reachable positions and speeds under second-order integrator dynamics with bounded acceleration in a given time $\tau$.

\paragraph{Safety constraints}
Using the previously-defined indicator variables $\varepsilon$ and the results from Section~\ref{seq:completed-coll}, we translate the safety constraints as follows: for every pair of conflicting robots $(i,j)$ and for all $0\leq k\leq K-1$:
\begin{align}
\pi_{ij}=1\wedge\varepsilon_{ij}^{\perp,out}(i,k)=0 \Rightarrow \varepsilon_{ij}^{\perp,in}(j,k+1) =& \ 0\tageq{s}\label{eq:cstr-safe-perp}\\
\pi_{ij}=1\wedge\varepsilon_{ij}^{\parallel,in}(j,k)=1  \wedge\varepsilon_{ij}^{\parallel,out}(i,k)=0 \Rightarrow & \nonumber \\ 
s_{i}^{k+1}-s_{j}^{k+1}& \geq a_{ij}\tageq{s}\label{eq:cstr-safe-pll}\\
\pi_{ij}=1\wedge\varepsilon_{ij}^{\parallel,in}(j,k)=1\wedge\varepsilon_{ij}^{\parallel,out}(i,k)=0\Rightarrow & \nonumber \\
s_{i}^{k+1}-s_{j}^{k+1}+\frac{\tau}{2}\left(v_{i}^{k+1}-v_{j}^{k+1}\right)& \geq  a_{ij} \tageq{s}\label{eq:cstr-safe-pll-speed}
\end{align}
with $a_{ij} = d_\parallel + \underline S_{ij}^{\parallel}(i)-\underline S_{ij}^{\perp}(j)$, in which $d_{\parallel}$ is a following distance
(from front of the follower to rear of the leader) between two robots. The term $\underline S_{ij}^{\parallel}(i)-\underline S_{ij}^{\perp}(j)$ in $a_{ij}$ accounts for the potential offset in curvilinear position between two robots in the case of merging paths (and vanishes in other cases), as illustrated in Fig.~\ref{fig:Completed-obstacle}. Condition
\eqref{eq:cstr-safe-perp} can be phrased as ``if i has priority and has not yet passed the collision set, then j cannot go in''. The additional condition (\ref{eq:cstr-safe-pll-speed}) is used to ensure that no collision occurs between two time steps; for the same reason, \eqref{eq:cstr-safe-perp} involves $\varepsilon_{ij}^{\perp,out}(i,{k})$ and $\varepsilon_{ij}^{\perp,in}(j,{k+1})$. This approach can be seen as a systematization of the concept of spatio-temporal trajectory envelopes presented in~\cite{Pecora2012,Cirillo2014}, where the temporal extent of these envelopes is automatically adjusted during resolution.
 
\setcounter{tagnumb}{0}
Note that the above formulation can be used in the case of a collision set with multiple connected components, by introducing a set of variables $\pi$ and $\varepsilon$, and of parameters $\underline{S}$ and $\overline{S}$ for each of these components. This method allows simultaneous resolution over multiple conflict areas, thus ensuring the solution is a global optimum and not comprised of several local optima, which would be the case if each connected component was treated separately.

\subsection{Optimization problem}
To simplify notations, we note $X$ the tuple of all the variables described above. The optimization problem (in which indices have been omitted for readability) of finding
\begin{align} \label{eq:optimization-problem}
\max_{X}\quad & \mathcal{O}(X) \\
  \small \mathrm{s.t.}\quad  \eqref{eq:cstr-mu-inside}-\eqref{eq:cstr-pij-xor},\ \eqref{eq:cstr-init-speed}-\eqref{eq:cstr-minmax-vel},\ & \small
 \eqref{eq:cstr-kinematic-inside}-\eqref{eq:cstr-accel-lb},\ \eqref{eq:cstr-safe-perp}-\eqref{eq:cstr-safe-pll-speed}\nonumber 
\end{align} 
either gives a solution to the discretized time-optimal coordination problem, or is infeasible. Specifically, infeasibilities can either be caused by the choice of a too small value for the number of time steps $K$, or because the initial states of the robots do not allow a safe passage through the intersection.

The sub-optimality caused by the time discretization vanishes as the time step goes to zero. Since the limit on computation time effectively sets a lower bound on the time step duration, it is desirable to choose a value providing good quality solutions in reasonable time. This issue will be discussed in Section~\ref{sec:Simulation-results}. However, an additional artifact arises from time discretization with a finite resolution, as objective function $\mathcal O$ does not distinguish solutions with sojourn times differing by less than the duration of a time step for at least one robot. To correct this issue, we make use of the fact that maximizing the speed of a robot allows to minimize its (non-discretized) sojourn time. Therefore, adding an ``averaged normalized speed'' term $\frac{1}{NK} \sum_{i,k} \frac{v_i^k}{\overline v_i}$ to function $\mathcal O$ allows to choose, among all solutions of~\eqref{eq:optimization-problem}, the one with highest average speed and thus smallest average sojourn time. Note that the weighting of the added term ensures this solution is still optimal for problem~\eqref{eq:optimization-problem}, and the modified objective function has been used in our simulations.

Moreover, our formulation could also be used for continuous arrivals of robots in real-time. Let $T$ be an upper bound for the computation time for $N_T$ robots: at a given time $t$ we consider the set $\mathcal{N}_{t}$ of robots entering the coordination region between times $t$ and $t+T$, and we assume $|\mathcal{N}_{t}| \leq N_T$. If optimal trajectories have been assigned for the robots of $\mathcal{N}_{t-T}$ and taking those as constraints for the robots of $\mathcal{N}_{t}$, we can compute optimal trajectories for those robots before they reach the entry of the coordination region. Constraint \eqref{eq:cstr-init-speed} ensures those trajectories remain feasible by that time and can therefore be assigned to the robots of $\mathcal{N}_t$. Note that this time-receding method may, however, cause sub-optimalities compared to considering all robots at once.

Also note that optimizing robots speed profiles for minimum sojourn time comes in pair with reducing safety margins to a minimum. For actual applications, this may cause problems in the case of unexpected events such as failure of a robot, which could make other robots unable to avoid a collision. Future work will study how to balance efficiency with the ability to cope with contingencies, to improve the robustness of the coordination.

\section{Simulation results\label{sec:Simulation-results}}
The use of the above optimization problem to find a time-optimal coordination has been validated by computer simulation on the example of autonomous vehicles in the intersection of Fig.~\ref{fig:Example-coord-region}. The simulation is based on the free traffic modeling tool SUMO \cite{SUMO2012} and uses its path generation algorithm to compute collisions sets. 

Vehicles are generated either deterministically (first simulation), or using random Poisson arrival times with normally-distributed entry speeds truncated to a minimum and a maximum speed (second and third simulations). Optimization problem \eqref{eq:optimization-problem} (using the modified objective function) is then run into the commercial MILP solver Gurobi~\cite{gurobi}, using its Python interface to generate the constraints. Lastly, if the problem is feasible, the solution trajectories are simulated in SUMO using the TraCI interface to verify that they do not generate collisions.

In all simulations, vehicles are modeled as rectangles of \SI{5}{\meter} length by \SI{2}{\meter} width, with $[\underline a_i, \overline a_i] = [-3, +4]$ \SI{}{\meter\per\second\squared}. The exit speed for all $i \in \mathcal{N}$ is set as  $v_i^{out} = \overline{v} = \SI{15}{\meter\per\second}$, which ensures the absence of collisions outside of the coordination region. The entry speed  $v_i^{in}$ is deterministically chosen in the first simulation and is normally distributed with average  \SI{12}{\meter\per\second} and standard deviation \SI{3}{\meter\per\second}, truncated to $[10,15]$ \SI{}{\meter\per\second} in the second and third simulations.

Simulations were performed on a personal computer running on a \SI{3.60}{\giga \hertz} Intel Core i7-4790 CPU with \SI{16}{\giga \byte} of RAM. A replay of some of our simulations is available in the accompanying video submission\footnote{Also available at \url{https://youtu.be/RiW2OFsdHOY}}.

\subsection{Microscopic simulation}
\begin{figure}
\subfloat[Optimal priorities\label{fig:opt-prio}]{\includegraphics[width=.5\columnwidth,height=3.5cm]{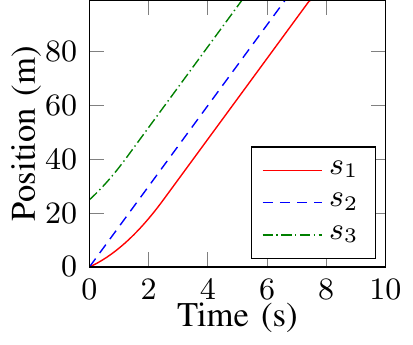} \hfill}
\subfloat[Suboptimal priorities\label{fig:subopt-prio}]{\includegraphics[width=.5\columnwidth,height=3.5cm]{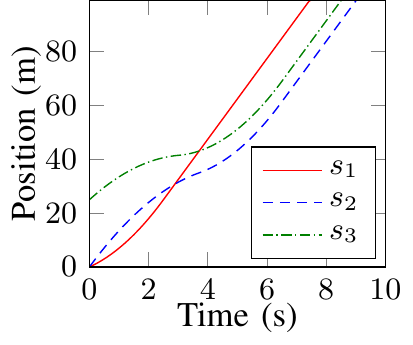}}
\caption{\label{fig:opt-trajectories}Optimal trajectories within given priority classes, corresponding to a global (left) and local (right) optimum}
\end{figure}

We first demonstrate the ability of our method to find the global optimum on a simple example with three vehicles in the case of the intersection displayed in Fig.~\ref{fig:Example-coord-region}: vehicle $1$ goes from south to west, vehicle $2$ from west to east and vehicle $3$ from north to south. Vehicles initially start at $(s_1^0, s_2^0, s_3^0) = (0,0,25)$ \SI{}{\meter} with speeds $(v_1^{in}, v_2^{in}, v_3^{in}) = (5, 15, 10)$ \SI{}{\meter\per\second}. Fig.~\ref{fig:opt-prio} shows the globally optimal trajectories for each vehicle, which lies in the homotopy class represented by priorities $3 \succ 2$, $2 \succ 1$, $3 \succ 1$. For comparison purposes, Fig.~\ref{fig:subopt-prio} shows the (locally) optimal trajectories when sub-optimal priorities $1 \succ 3$, $3 \succ 2$, $1 \succ 2$ are enforced. The optimum average sojourn time found by our algorithm is \SI{6.5}{s}, and the example sub-optimal one is \SI{8.4}{s}.

\subsection{Influence of time step duration}
The discretization time step has a double effect on the solution: first, we assume constant acceleration during one time step. Second, the safety constraints require that one robot of each conflicting pair leaves the conflict area one time step before the other can enter.

\begin{figure}
\centering \includegraphics[width=\columnwidth,height=4cm]{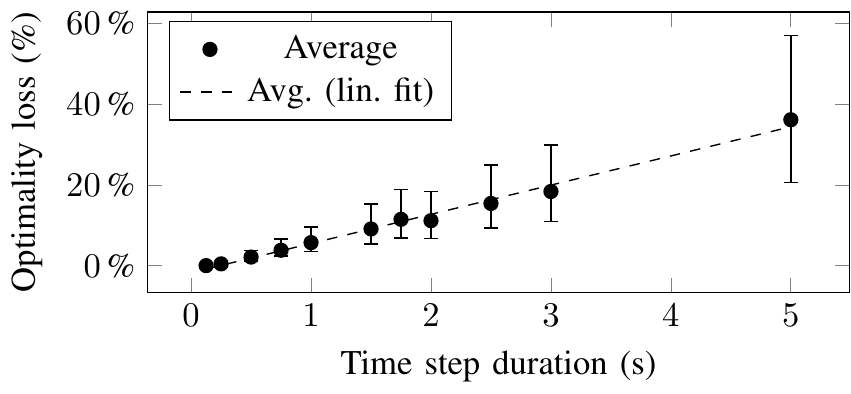}
\caption{\label{fig:timestep-influence} Average relative optimality loss (compared to a \SI{0.125}{\second} time step), depending on time step duration, for $85$ instances of $15$ vehicles. Error bars correspond to $1$ standard deviation for instances above or below average.\vspace*{-0.5cm}}
\end{figure}

In Fig.~\ref{fig:timestep-influence} we show the average optimality loss caused by choosing larger time step durations for a random set of $85$ initial configurations of $15$ vehicles. For each instance, the resolution-optimal average time $t_{opt}^{\tau}$ is computed for time step durations $\tau$ ranging from \SI{0.125}{\second} to \SI{5}{\second}. The relative loss of optimality is computed as $\frac{t_{opt}^{\tau} - t_{opt}^{0.125}}{t_{opt}^{0.125}}$. Interestingly, the averaged values fit closely to an affine function with slope $7.2$\% per second for the above set of parameters. The loss of optimality remains less than $6$\% when the time step is smaller than \SI{1}{\second}; moreover, the solution of \eqref{eq:optimization-problem} converges as the time step duration vanishes.

\begin{figure}
\centering \includegraphics[width=\columnwidth,height=4cm]{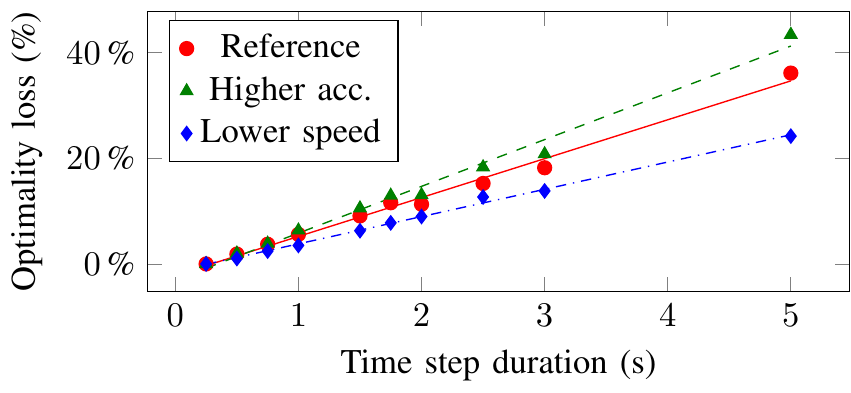}
\caption{\label{fig:timestep-slope} Average relative optimality loss (compared to a \SI{0.25}{\second} time step), depending on time step duration, for $85$ instances of $15$ vehicles with reference parameters (red), lower maximum speed of \SI{10}{\meter \per \second} (blue) and higher absolute values of accelerations $[\underline a_i, \overline a_i] = [-6, +8]$ \SI{}{\meter\per\second\squared} (green).}
\end{figure}

The kinodynamic parameters, \textit{i.e.} the maximum speed $\overline{v}$ and the acceleration bounds $\underline a_i$ and $\overline a_i$ influence the magnitude of the optimality loss. Fig.~\ref{fig:timestep-slope} shows a comparison of the losses of optimality for three different scenarios, namely ``reference'', ``lower speed'' and ``higher acceleration''. Reference parameters are, as before, $\left( \overline{v}, \underline a_i, \overline a_i \right) = (\SI{15}{\meter\per\second}, \SI{-3}{\meter\per\second\squared}, \SI{+4}{\meter\per\second\squared})$. The same set of $85$ initial configurations is used for those three scenarios, and we only vary the kinodynamic parameters; in the lower speed scenario, initial speeds are also reduced by $33$\%. We find that an increase of time step duration causes higher losses of optimality in instances with more dynamic robots (higher speeds or higher absolute values of acceleration bounds) than those with less dynamic ones. As a result, the time step duration can be adapted to the dynamic characteristics of robots; for instance using longer time steps for slower robots.

\subsection{Computation time}

\begin{figure}
	\centering \includegraphics[width=\columnwidth,height=4cm]{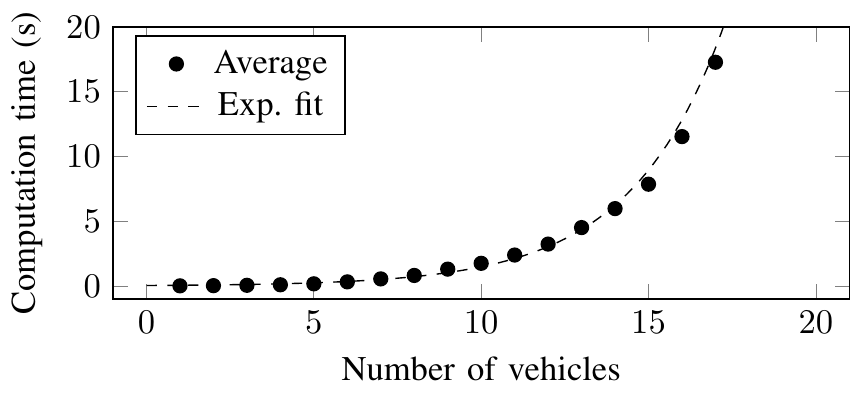}
	\caption{\label{fig:comput-time-num}Averaged computation time over 10 instances for a time step duration of \SI{1}{\second} and a varying number of vehicles.\vspace*{-0.5cm}}
\end{figure}
To measure the computation time required to solve problem~\eqref{eq:optimization-problem} for the intersection displayed in Fig.~\ref{fig:Example-coord-region}, we run the simulator for different numbers of vehicles with a fixed time step of \SI{1}{\second} (Fig.~\ref{fig:comput-time-num}). Computation time remains below \SI{1}{\second} for up to 8 vehicles, which would make our approach suitable for real-world applications. Note that the same \SI{30}{\second} time horizon is considered across all simulations to provide fair comparison, while a shorter horizon could be used for problems involving fewer robots, thus reducing computation time; moreover, the presented MILP problem has been formulated for clarity rather than execution speed, notably by introducing redundant variables. As a result, performance can likely be further improved to process more robots in the same time frame.

\section{Conclusion\label{sec:Conclusion}} 

This article presents a new approach to compute time-optimal trajectories for the coordination of mobile robots in a structured environment, taking into account kinodynamic constraints. Assuming each robot follows a predetermined path and using well-chosen bounding polygons, it is possible to formulate this problem as a mixed-integer linear program. By using temporal instead of spatial discretization, our approach allows an exact modeling of second-order integrator dynamics with piecewise-constant acceleration. The formulated problem mixes the discrete choice of relative priorities between conflicting robots with the continuous optimization of speed profiles respecting these priorities.

Extensive computer simulations have been run, on the example of  a road intersection with autonomous vehicles, using the open-source traffic simulator SUMO and commercial MILP solver GUROBI. These simulations demonstrate the ability of our approach to compute optimal, collision-free trajectories in reasonable time.

Using consumer-grade hardware, the proposed method can treat up to 8 robots in less than a second. It is therefore suitable for real-time use, for instance for robots in an automated warehouse or autonomous vehicles at an intersection. In situations where computation time is more important than optimality, our formulation could also be used to design more efficient heuristics, with provable performance bounds, or to assess potentially dangerous situations where no safe trajectories exist. More general quadratic programming techniques could allow to use more complex objective functions, and relax the constant acceleration assumption while still providing an exact modeling of the dynamic constraints, opening many perspectives for future research.

\bibliographystyle{IEEEtranurl}
\bibliography{iros2016}

\end{document}